# Practical Methods for Proving Termination of General Logic Programs

**Elena Marchiori**                                                    ELENA@CWI.NL
*Centrum voor Wiskunde en Informatica (CWI)*
*P.O. Box 94079, 1090 GB Amsterdam, The Netherlands*

## Abstract

Termination of logic programs with negated body atoms (here called general logic programs) is an important topic. One reason is that many computational mechanisms used to process negated atoms, like Clark's negation as failure and Chan's constructive negation, are based on termination conditions. This paper introduces a methodology for proving termination of general logic programs w.r.t. the Prolog selection rule. The idea is to distinguish parts of the program depending on whether or not their termination depends on the selection rule. To this end, the notions of low-, weakly up-, and up-acceptable program are introduced. We use these notions to develop a methodology for proving termination of general logic programs, and show how interesting problems in non-monotonic reasoning can be formalized and implemented by means of terminating general logic programs.

## 1. Introduction

General logic programs (GLP's for short) provide formalizations and implementations for special forms of non-monotonic reasoning, as illustrated by Apt and Bol (1994) and Baral and Gelfond (1994). For example, Prolog's negation as finite failure operator can be used to implement the temporal persistence problem in Artificial Intelligence as a logic program (Kowalski & Sergot, 1986; Evans, 1990; Apt & Bezem, 1991). The implementation of operators like Clark's negation as failure (Clark, 1978) and Chan's constructive negation (Chan, 1988), is based on termination conditions. Therefore the study of termination of GLP's (e.g., De Schreye & Decorte, 1994) is an important topic.

Two classes of GLP's that behave well w.r.t. termination are the so-called acyclic and acceptable programs (Apt & Bezem, 1991; Apt & Pedreschi, 1991). In fact, Apt and Bezem (1991) prove that if negation as finite failure is incorporated into the proof theory, then for any acyclic program, all sld-derivations with arbitrary selection rule of ground queries terminate. The converse of this result, i.e., if a program terminates for all ground queries, then it is acyclic, holds only under the assumption that the program is 'non-floundering'. Apt and Pedreschi (1991) establish analogous results on termination for so-called acceptable programs, this time w.r.t. the Prolog selection rule, which selects the leftmost literal of a query.

Floundering is an abnormal form of termination which arises as soon as a non-ground negated atom is selected, as explained e.g., in (Apt & Bol, 1994). To treat also non-ground negated atoms, Chan (1988) introduced a procedure known as Chan's constructive negation. Using Chan's constructive negation, Marchiori (1996) showed that the notions of acyclicity and acceptability provide a complete characterization of programs that terminate for all ground queries.

The notion of acceptability combines the definition of acyclicity with a semantic condition, and therefore proving acceptability may be rather cumbersome. The aim of this paper is to develop a methodology for proving termination with respect to the Prolog selection rule, by using as little semantic information as possible. A program $P$ is split into two parts, say $P_1$ and $P_2$; then one part is proven to be acyclic, the other one to be acceptable, and these results are combined to conclude that the original program is terminating w.r.t. the Prolog selection rule. The decomposition of $P$ is done in such a way that no relations defined in $P_1$ occur in $P_2$. We introduce the notions of *up-acceptability*, where $P_1$ is proven to be acceptable and $P_2$ to be acyclic, and of *low-acceptability*, which treats the converse case ($P_1$ acyclic and $P_2$ acceptable). In order to be of more practical use, the notion of up-acceptability is generalized to *weak up-acceptability*. We integrate these notions in a bottom-up methodology for proving termination of general logic programs. We apply our results to programs formalizing problems in non-monotonic reasoning. In particular, we show that the *planning in the blocks world* problem can be formalized and implemented by means of an up-acceptable program. This provides a class of queries (up-bounded queries) that can be completely answered.

Even though our main theorems (Theorem 5.5, 6.4 and 7.2) deal with Chan's constructive negation only, a simple inspection of the proofs shows that they hold equally well for the case of negation as finite failure.

Our approach provides a simple methodology for proving termination of GLP's, by combining the results of Bezem, Apt and Pedreschi on acyclic and acceptable programs. The relevance of this methodology is twofold: for a large class of programs, it overcomes the drawback of the method of Apt and Pedreschi (1991), namely the use of too much semantic information; and it allows to identify those parts of the program whose termination is dependent on the use of the Prolog selection rule. Moreover, the examples that are given, show that systems based on the logic programming paradigm provide a suitable formalization and implementation for problems in non-monotonic reasoning.

The paper is organized as follows. The next section contains some terminology and preliminaries. In Sections 3 and 4 the notions of acyclicity and acceptability are presented. Sections 5, 6, and 7, contain our alternative definitions of acceptability. In Section 8 these definitions are integrated in a methodology for proving termination. Finally, in Section 9 some conclusions are given. This paper is an extended and revised version of (Marchiori, 1995).

## 2. Preliminaries

The following notation will be used. We follow Prolog syntax and assume that a string starting with a capital letter represents a variable, while other strings represent constants, terms and relations. Relation symbols are often denoted by $p, q, r$. A literal is either an atom $p(s_1, \ldots, s_k)$, or a negated atom $\neg p(s_1, \ldots, s_k)$, or an equality $s = t$, or an inequality $\forall(s \neq t)$, where $\forall$ quantifies over some (perhaps none) of the variables occurring in $s$, $t$. Equalities and inequalities are also called *constraints*, and denoted by $c$. An inequality $\forall(s \neq t)$ is *primitive* if it is satisfiable but not valid. For instance, $X \neq a$ is primitive. An (extended) general logic *program*, denoted by $P$, $R$, is a finite set of clauses

$$H \leftarrow L_1, \ldots, L_m.$$

with $m \geq 0$, where $H$ is an atom, and $L_i$ is a literal, for $i \in [1, m]$. A *query* is a finite sequence of literals, and is denoted by $Q$.

To treat negated non-ground atoms, Chan (1988) proposes to augment `sld`-resolution with a procedure, informally described as follows. For a substitution $\theta = \{X_1/t_1, \ldots, X_n/t_n\}$, we denote by $E_\theta$ the equality formula $(X_1 = t_1 \wedge \ldots \wedge X_n = t_n)$. For any negated atom $\neg A$, if all the `sld`-derivations of $A$ are finite, and $\theta_1, \ldots, \theta_k$, with $k \geq 0$, are the computed answer substitutions, then the answers for $\neg A$ are obtained from the negation of $\exists (E_{\theta_1} \vee \ldots \vee E_{\theta_k})$, where $\exists$ quantifies over the variables not occurring in $A$. For instance, consider the program

```
p(a) ←.
p(b) ←.
```

The answer to the query $\neg p(X)$ is $X \neq a \wedge X \neq b$. We call `sldcnf`-resolution, `sld`-resolution augmented with Chan's procedure. To show the correctness of `sldcnf`-resolution, we choose as program semantics the Clark's completion (Clark, 1978). This semantics is a natural interpretation of a GLP as a set of definitions. Intuitively, the Clark's completion of a program $P$, denoted by $comp(P)$, is the first-order theory obtained by replacing the implication of each clause of $P$ with an equivalence. It is constructed as follows. Below, $\forall$ quantifies over $X_1, \ldots, X_k$.

- For every relation symbol $p$ occurring in $P$, having say $k \geq 0$ arguments:

  - if $p$ does not occur in the head of any clause then add the formula
    $\forall (p(X_1, \ldots, X_k) \leftrightarrow \textit{false})$;

  - otherwise, if $k = 0$ then add the formula $p \leftrightarrow \textit{true}$; if $k > 0$ and $C_1, \ldots, C_l$, with $l \geq 1$ are all the clauses of $P$ with head symbol $p$, with $C_i = p(s_1^i, \ldots, s_k^i) \leftarrow Q_i$, then add the formula $\forall (p(X_1, \ldots, X_k) \leftrightarrow \vee_{i \in [1,l]} (\exists \mathcal{V}_i (E_i \wedge Q_i)))$, where $\mathcal{V}_i$ is the set of variables of $C_i$, $E_i$ is $(s_1^i = X_1 \wedge \ldots \wedge s_k^i = X_k)$, and $X_1, \ldots, X_k$ are fresh variables.

- Finally, the following *free equality axioms* are added, so that the equality theory of $comp(P)$ becomes the same as the one of the Herbrand universe.

  - $f(X_1, \ldots, X_k) = f(Y_1, \ldots, Y_k) \rightarrow (X_1 = Y_1 \wedge \ldots \wedge X_k = Y_k)$,
    for every function symbol $f$,

  - $f(X_1, \ldots, X_k) \neq g(Y_1, \ldots, Y_m)$,
    for every distinct function symbols $f$ and $g$,

  - $X \neq s$,
    for every term $s$ s.t. $X$ occurs in $s$.

The soundness of `sldcnf`-resolution w.r.t. Clark's semantics follows from

$$comp(P) \models \forall (A \leftrightarrow \exists (E_{\theta_1} \vee \ldots \vee E_{\theta_k})),$$

where $\forall$ quantifies over all the free variables of the formula. `sldcnf`-resolution is complete only for queries having all terminating derivation. In fact, Chan's procedure is not defined if $A$ has an infinite derivation. As a consequence, the notion of (infinite) derivation is not always defined. This is a problem for the study of termination of GLP's, because the notion of derivation is of crucial importance. Therefore, we refer here to an alternative definition of Chan's procedure given by Marchiori (1996), where the subtrees used to resolve negative literals are built in a top-down way, constructing their branches in parallel. As a consequence, the main derivation is infinite if at least one of these subtrees is infinite.

Termination of GLP's depends on the selection rule. For instance, the program

```
p ← q,p.
```

terminates if the Prolog selection rule, which chooses the leftmost literal of a query, is used. But, the program does not terminate if the selection rule which chooses the rightmost literal of a query is used. We shall consider the generalization of the Prolog selection rule to programs containing constraints, which delays the selection of primitive constraints as follows: the leftmost literal of a query which is not a primitive inequality is chosen. For simplicity, we continue to refer to this selection rule as the *Prolog selection rule*. An `sldcnf`-tree that is obtained by using the Prolog selection rule is called `ldcnf`-tree.

To prove termination of logic programs, suitable functions from ground atoms to natural numbers, called level mappings, will be used. Let $\mathcal{B}_P$ denote the Herbrand base of $P$.

**Definition 2.1 (Level Mapping)** A *level mapping* (for $P$) is a function $|\ |$ from $\mathcal{B}_P$ to natural numbers.                                                                                   □

A level mapping is extended to negated ground atoms by $|\neg A| = |A|$. We do not need to extend this notion also to constraints, because they represent terminating atomic actions. However, note that the presence of constraints in a query influences termination, because, for instance, a derivation finitely fails if an unsatisfiable constraint is selected.

## 3. Acyclic Programs

Our method will be based on the notions of acyclicity and acceptability, which are used to characterize a class of terminating programs w.r.t. an arbitrary and the Prolog selection rule, respectively. In this section we recall the definition of acyclicity, and some useful results from (Marchiori, 1996), while acceptability will be discussed in Section 4.

Apt and Bezem (1991) study termination of GLP's w.r.t. an arbitrary selection rule. They introduce the following elegant syntactic notion.

**Definition 3.1 (Acyclic Program)** A program $P$ is *acyclic w.r.t. a level mapping* $|\ |$ if for all ground instances $H \leftarrow L_1, \ldots, L_n$ of clauses of $P$ we have that $|H| > |L_i|$ holds for all $i \leq 1 \leq n$ s.t. $L_i$ is not a constraint. $P$ is *acyclic* if there exists a level mapping $|\ |$ s.t. $P$ is acyclic w.r.t. $|\ |$.                                                                      □

If a program is acyclic, then all ground queries have only finite derivations, and hence terminate. To extend this result to non-ground queries, the following notion of boundedness is used.

**Definition 3.2 (Bounded Query)** Let $|\ |$ be a level mapping. A query $Q = L_1, \ldots, L_n$ is bounded (w.r.t. $|\ |$) if for every $1 \leq i \leq n$, the set

$$|Q|_i = \{|L_i'| \mid L_i' \text{ is a ground instance of } L_i\}$$

is finite.                                                                                      □

Notice that ground queries are bounded. Apt and Bezem prove that for an acyclic program, every bounded query $Q$ has only finite derivations w.r.t. negation as finite failure. The converse of this result does not hold, due to the possibility of floundering. Instead, using Chan's constructive negation, we obtain a complete characterization (Marchiori, 1996).

First, we formalize the concept of termination w.r.t. an arbitrary selection rule.

**Definition 3.3 (Terminating Query and Program)** A query is *terminating* (w.r.t. $P$) if all its `sldcnf`-derivations (in $P$) are finite. A program $P$ is *terminating* if all *ground* queries are terminating w.r.t. $P$.                                                                □

**Theorem 3.4** *Let $P$ be an acyclic program and let $Q$ be a bounded query. Then every* `sldcnf`-*tree for $Q$ in $P$ contains only bounded queries and is finite.*

**Theorem 3.5** *Let $P$ be a terminating program. Then there exists a level mapping $| \ |$ s.t.:* **(i)** *$P$ is acyclic w.r.t. $| \ |$;* **(ii)** *for every query $Q$, $Q$ is bounded w.r.t. $| \ |$ iff $Q$ is terminating.*

From Theorems 3.4 and 3.5 it follows that terminating programs coincide with acyclic programs and that for acyclic programs a query has a finite `sldcnf`-tree if and only if it is bounded. Notice that when negation as finite failure is assumed, Theorem 3.5 does not hold. For instance, the program:

```
p(X) ← ¬ q(Y).
q(s(X)) ← q(X).
q(0) ←.
```

is terminating (floundering) but it is not acyclic.

Finding a level mapping for proving acyclicity is a creative process. We refer the reader to (De Schreye & Decorte, 1994) for a thorough presentation of various techniques for constructing level mappings.

The following section illustrates how an interesting problem in nonmonotonic reasoning can be formalized and implemented as an acyclic program.

## 3.1 An Example: Blocks World

The blocks world is a formulation of a problem in AI, where a robot performs a number of primitive actions in a simple world (see for instance Nilsson, 1982). Here we consider a simpler version of this problem by Sacerdoti (1977). There are three blocks $a$, $b$, $c$, and three different positions $p$, $q$, $r$ on a table. A block can lay either above another block or on one of these positions, and it can be moved from one position to another. The problem consists of specifying possible configurations, i.e., those obtained from the initial situation by performing a sequence of possible moves. An example of an initial situation is given in Figure 1.

Kowalski (1979) gives a clausal representation of this problem by means of pre- and post-conditions. Here we formulate the problem using McCarthy and Hayes' situation calculus

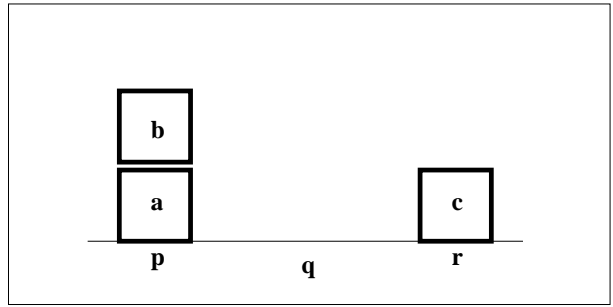

Figure 1: The Blocks-World

(McCarthy & Hayes, 1969), in terms of facts, events and situations. There are three types of *facts*: $loc(X, L)$ stands for 'block $X$ is in location $L$'; $above(X, Y)$ for 'block $X$ is on block $Y$'; and $clear(L)$ for 'there is no block in location $L$'. There is only one type of *event*: $move(X, L)$ stands for 'move block $X$ into location $L$'. Finally, *situations* are described using lists: [ ] denotes the initial situation, and $[Xe|Xs]$ the situation obtained from $Xs$ by performing the event $Xe$. Based on the above representation, the blocks world can be formalized as the following GLP BLOCKSWORLD:

1)  `holds(l,[])` $\leftarrow$ . `l` $\in \mathcal{L}$
2)  `block(bl)` $\leftarrow$ . `bl` $\in \mathcal{B}$
3)  `position(pl)` $\leftarrow$ . `pl` $\in \mathcal{P}$
4)  `holds(loc(X,L),[move(X,L)|Xs])` $\leftarrow$
     `block(X),`
     `position(L),`
     `holds(clear(top(X)),Xs),`
     `holds(clear(L),Xs),`
     `L` $\neq$ `top(X).`
5)  `holds(loc(X,L),[Xe|Xs])` $\leftarrow$
     `block(X),`
     `position(L),`
     $\neg$ `abnormal(loc(X,L),Xe,Xs),`
     `holds(loc(X,L),Xs).`
6)  `holds(above(X,Y),Xs)` $\leftarrow$
     `holds(loc(X,top(Y)),Xs).`
7)  `holds(above(X,Y),Xs)` $\leftarrow$
     `holds(loc(X,top(Z)),Xs),`
     `holds(loc(Z,top(Y)),Xs).`
8)  `holds(clear(L),Xs)` $\leftarrow$
     $\neg$ `occupied(L,Xs).`
9)  `abnormal(loc(X,L),move(X,L'),Xs)` $\leftarrow$.

```
10) occupied(L,Xs)←
        holds(loc(X,L),Xs).
11) legals([(a,L1),(b,L2),(c,L3)],Xs) ←
        holds(loc(a,L1),Xs),
        holds(loc(b,L2),Xs),
        holds(loc(c,L3),Xs).
```

Here $\textbf{top(X)}$ denotes the top of block $\textbf{X}$, $\mathcal{B} = \{a, b, c\}$, $\mathcal{P} = \{p, q, r, top(a), top(b), top(c)\}$, and $\mathcal{L} = \{loc(a, p), loc(b, q), loc(c, r)\}$. Moreover, lines 1, 2 and 3 abbreviate sets of clauses, and line 1 specifies the initial situation. The relation $\textbf{holds}$ describes when a fact is possible in a given situation, and the relation $\textbf{legals}$ when a configuration is possible in a given situation.

Consider the following level mapping, where for a ground term $y$, $|y|$ denotes the length of the list $y$, otherwise (i.e., if $y$ is not a list) $|y|$ is 0.

$$|block(x)| = 0,$$
$$|position(x)| = 0,$$
$$|abnormal(x, y, z)| = 0,$$

$$|holds(x, y)| = \left\{ \begin{array}{ll} 3 * |y| + 1 & \text{if } x \text{ is of the form } loc(r, s), \\ 3 * |y| + 3 & \text{if } x \text{ is of the form } clear(r, s), \\ 3 * |y| + 4 & \text{if } x \text{ is of the form } above(r, s), \\ 0 & \text{otherwise.} \end{array} \right.$$

$$|occupied(x, y)| = 3 * |y| + 2,$$
$$|legals(x, y)| = 3 * |y| + 2.$$

It is easy to check that BLOCKSWORLD is acyclic w.r.t. $|\,|$.

Therefore, the class of questions expressed by means of bounded queries can be completely answered. For instance, the question 'when block $a$ remains in its initial position $p$ under the occurrence of an action?' can be formalized as the query $\textbf{holds(loc(a,p),[A])}$. This query is bounded, hence every of its $\textbf{sldcnf}$-derivations is finite, with answer $\forall L(A \neq move(a, L))$.

Note that this query would flounder when negation as finite failure is used.

## 4. Acceptable Programs

In the previous section, we have seen how termination of GLP's w.r.t. an arbitrary selection rule can be proven by means of the notion of acyclicity. The notion of acceptability (Apt & Pedreschi, 1991) is used for proving termination of GLP's w.r.t. the Prolog selection rule. In this section, we recall this notion, together with some useful results from (Marchiori, 1996). Acyclicity and acceptability will be combined in the following sections to provide more practical tools for proving termination of GLP's w.r.t. the Prolog selection rule.

In order to study termination of general logic programs with respect to the Prolog selection rule, Apt and Pedreschi (1991) introduced the notion of acceptable program. This notion is based on the same condition used to define acyclic programs, except that, for a ground instance $H \leftarrow L_1, \ldots, L_n$ of a clause, the test $|H| > |L_i|$ is performed only until the first literal $L_{\overline{n}}$ which fails. This is sufficient since, due to the Prolog selection rule, literals after $L_{\overline{n}}$ will not be selected. To compute $\overline{n}$, a class of models of $P$, here called *specialized models*, is used. The following notion is used. The *restriction* of an interpretation $I$ to a set $S$ of relations, denoted by $I_{|S}$, is the set of atoms of $I$ having their relations in $S$.

**Definition 4.1 (Specialized Model)** Let $Neg_P$ be the least set $S$ of relations s.t.: the relations of $P$ occurring in negated atoms are in $S$; and if an element of $S$ occurs in the head of a clause, then all the relations occurring in the body of that clause are in $S$. Let $P^-$ be the set of clauses in $P$ whose head contains a relation from $Neg_P$. Now a model $I$ of $P$ is *specialized* if $I_{|Neg_P}$ is a model of $comp(P^-)$. $\square$

**Definition 4.2 (Acceptable Program)** Let $|\ |$ be a level mapping for $P$ and let $I$ be an interpretation of $P$. $P$ is *acceptable w.r.t.* $|\ |$ *and* $I$ if $I$ is a specialized model of $P$, and for all ground instances $H \leftarrow L_1, \ldots, L_n$ of clauses of $P$ we have that $|H| > |L_i|$ holds for every $1 \leq i \leq \overline{n}$ s.t. $L_i$ is not a constraint, where $\overline{n} = min(\{n\} \cup \{i \in [1, n] \mid I \not\models L_i\})$. $P$ is *acceptable* if it is acceptable w.r.t. some level mapping and interpretation. $\square$

If a program is acceptable, then every ground query has only finite `ldcnf`-derivations, hence it terminates. To extend this result to non-ground queries, as for the acyclic case, the following notion of boundedness is used.

**Definition 4.3 (Bounded Query)** Let $|\ |$ be a level mapping and let $I$ be a specialized model of $P$. A query $Q = L_1, \ldots, L_n$ is bounded (w.r.t. $|\ |$ and $I$) if for every $1 \leq i \leq n$

$$|Q|_I^i = \{|L_i'| \mid \quad L_1', \ldots, L_i' \text{ ground instance of } L_1, \ldots, L_i \text{ and} \\ I \models L_1', \ldots, L_{i-1}'\}$$

is finite. $\square$

Apt and Pedreschi prove that for an acceptable program, every bounded query has only finite derivations w.r.t. the Prolog selection rule and negation as finite failure. The converse of this result holds when Chan's constructive negation is used (Marchiori, 1996). First, we formalize the concept of termination w.r.t. the Prolog selection rule.

**Definition 4.4 (Left-Terminating Query and Program)** A query is *left-terminating* (w.r.t. $P$) if all its `ldcnf`-derivations are finite. A program $P$ is *left-terminating* if every ground query is left-terminating w.r.t. $P$. $\square$

**Theorem 4.5** *Let $P$ be an acceptable program and let $Q$ be a bounded query. Then every* `ldcnf`*-tree for $Q$ in $P$ contains only bounded queries and is finite.*

**Theorem 4.6** *Let $P$ be a left-terminating program. Then there exists a level mapping $|\ |$, and a specialized model $I$ of $P$ s.t.:* **(i)** *$P$ is acceptable w.r.t. $|\ |$ and $I$;* **(ii)** *for every query $Q$, $Q$ is bounded w.r.t. $|\ |$ and $I$ iff $Q$ is left-terminating.*

In the following section an acceptable program that formalizes planning in the blocks world is given.

## 4.1 An Example: Planning in the Blocks World

Consider planning in the blocks world, amounting to the specification of a sequence of possible moves transforming the initial configuration into a final configuration, e.g., as in Figure 2. This problem can be solved using a nondeterministic algorithm (Sterling & Shapiro, 1994): *while the desired configuration has not yet been reached, find a legal action, update the current configuration, and check that it was not already obtained.* The following program PLANNING follows this approach: it consists of all the clauses of the program BLOCKSWORLD, minus 6) and 7), and plus the following clauses:

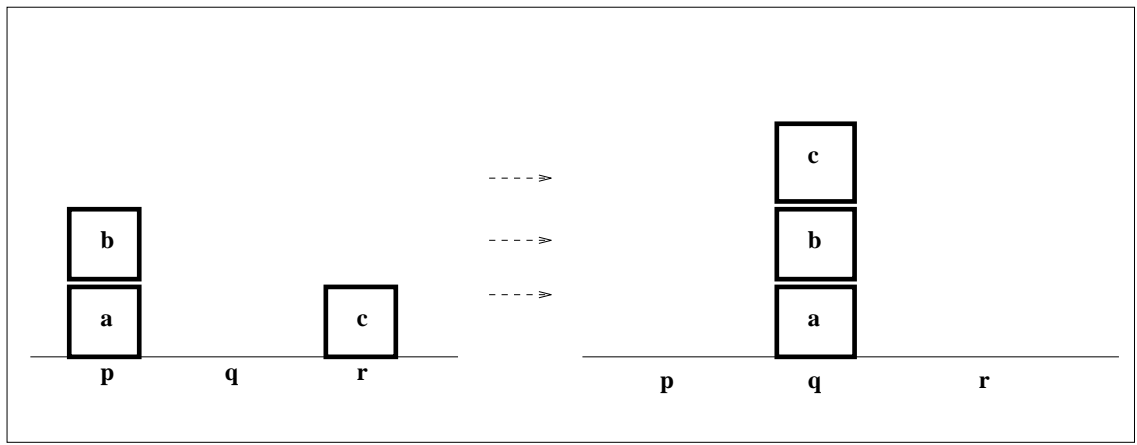

Figure 2: Planning in the Blocks-World

$1p)$   `transform(Xs,St,Plan) ←`
    `state(St0),`
    `legals(St0,Xs),`
    `trans(Xs,St,[St0],Plan).`
$2p)$   `trans(Xs,St,Vis,[ ]) ←`
    `legals(St,Xs).`
$3p)$   `trans(Xs,St,Vis,[Act|Acts]) ←`
    `state(St1),`
    `¬ member(St1,Vis),`
    `legals(St1,[Act|Xs]),`
    `trans([Act|Xs],St,[St1|Vis],Acts).`
$4p)$   `state([(a,L1),(b,L2),(c,L3)]) ←`
    `P=[p,q,r,top(a),top(b),top(c)],`
    `member(L1,P),`
    `member(L2,P),`
    `member(L3,P).`
$5p)$   `member(X,[X|Y]) ←.`

```
6p)  member(X,[Y|Z]) ←
         member(X,Z).
```

Planning in the blocks-world is specified by the relation **transform**: in clause $1p$) first a legal configuration for the actual situation is found by means of the predicate **legals**; then the predicate **trans** is used to construct incrementally a plan from this configuration to the final one. It uses an accumulator as third argument, to guarantee that a plan does not pass twice through the same configuration. Clause $3p$) takes care of expanding a plan: it first looks for a configuration which was not already considered, and then it adds to the plan the legal action yielding that configuration. Clause $2p$) guarantees termination of the construction when the final configuration is reached.

To prove the acceptability of PLANNING, we have to find a model of PLANNING that is also a model of $comp(\{5p),6p)\} \cup$ BLOCKSWORLD$\setminus\{6),7),11)\})$. We do not need to use all this semantic information, because from the acyclicity of BLOCKSWORLD, it follows that PLANNING is left-terminating if the following program TRAS is acceptable. We postpone the justification of this claim till the next section.

```
1'p)  transform(Xs,St,Plan) ←
          state(St0),
          trans(Xs,St,[St0],Plan).
2p)   trans(Xs,St,Vis,[ ]) ←.
3'p)  trans(Xs,St,Vis,[Act|Acts]) ←
          state(St1),
          ¬ member(St1,Vis),
          trans([Act|Xs],St,[St1|Vis],Acts).
4p)   state([(a,L1),(b,L2),(c,L3)]) ←
          P=[p,q,r,top(a),top(b),top(c)],
          member(L1,P),
          member(L2,P),
          member(L3,P).
5p)   member(X,[X|Y]) ←.
6p)   member(X,[Y|Z]) ←
          member(X,Z).
```

TRAS is obtained from PLANNING by first deleting the subprogram 'defining' **legals**, and next the literals with relation **legals** occurring in the body of the remaining clauses. By considering TRAS, we need less semantic information, namely a model of TRAS that is also a model of $comp(\{5p),6)\})$. To show that TRAS is acceptable, we consider the following level mapping:

$|member(x,y)| = |y|;$
$|state(x)| = 7;$
$|trans(x,y,z,w)| = tot - card(el(z) \cap S) + 3 * (|x| + 1) + 5 + |z|;$

$$|transform(x, y, z)| = tot + 3 * (|x| + 1) + 6.$$

Above, $S$ denotes $\{[(a, p1), (b, p2), (c, p3)] \mid \{p1, p2, p3\} \subset \{p, q, r, top(a), top(b), top(c)\}\}$, and $tot$ is the cardinality of $S$. Moreover, if $z$ is a list then $el(z)$ denotes the set of its elements, otherwise it denotes the empty set; $card(el(z) \cap S)$ is the cardinality of the set $el(z) \cap S$; finally, if $x$ is a list then $|x|$ denotes its length, otherwise it denotes 0. Observe that $(tot - card(el(z) \cap S)) \geq 0$. Thus $|\ |$ is well defined. For an atom $p(s_1, \ldots, s_n)$, we denote by $[p(s_1, \ldots, s_n)]$ the set of all its ground instances. Consider the following interpretation $I = I_{transform} \cup I_{trans} \cup I_{member} \cup I_{state}$ of TRAS, with:

$I_{transform} = [transform(X, Y, Z)],$
$I_{trans} = [trans(X, Y, Z, W)],$
$I_{member} = \{member(x, y) \mid y \text{ is a list s.t. } x \in set(y)\},$
$I_{state} = \{state(x) \mid x \in S\}.$

It is easy to prove that $I$ is a model of TRAS. Moreover, $Neg_{tras} = \{member\}$, and TRAS$^-$ is equal to $\{5p), 6p)\}$. So, $I_{|\{member\}}$ is a model of $comp(\text{TRAS}^-)$. To show that TRAS is acceptable w.r.t. $I$ and $|\ |$, we use the following properties of $|\ |$, which are readily verified:

$$|transform(x, y, z)|_1 \geq 8, \tag{1}$$

$$|trans(x, y, z, w)|_1 \geq 8, \tag{2}$$

$$|trans(x, y, z, w)|_1 > |z|. \tag{3}$$

The proof of the acceptability of TRAS proceeds as follows:

- Consider a ground instance:
  $transform(xs, xt, plan) \leftarrow state(st0), trans(xs, st, [st0], plan).$
  of $1p)$. From (1) it follows that:

$$|transform(xs, xt, plan)| > |state(st0)|.$$

  Suppose that $I \models state(st0)$. Then $st0 \in S$, so $card(el(S \cap el([st0]))) = 1$; hence:

$$|transform(xs, xt, plan)| > |trans(xs, st, [st0], plan)|.$$

- Consider a ground instance:
  $trans(xs, st, vis, [act|acts]) \leftarrow$
  $\qquad state(st1), \neg member(st1, vis), trans([act|xs], st, [st1|vis], acts).$
  of $2'p)$. From (2) it follows that:

$$|trans(xs, st, vis, [act|acts])| > |state(st1)|,$$

and from (3):

$$|trans(xs, st, vis, [act|acts])| > |\neg member(st1, vis)|.$$

Suppose that $I \models state(st1), \neg member(st1, vis)$. Then $st1 \in S$, but $st1 \notin set(vis)$; so $card(S \cap el([st1|vis])) = card(S \cap el(vis)) + 1$; hence $tot - card(S \cap el([st1|vis])) < tot - card(S \cap el(vis))$. Therefore,

$$|trans(xs, st, vis, [act|acts])| > |trans([act|xs], st, [st1|vis], acts)|.$$

- The proof for the other clauses of TRAS is similar.

## 5. Up-Acceptability

In this section, we introduce a first integration, called up-acceptability, of the notions of acyclicity and acceptability. We show that up-acceptability provides a more practical tool than acceptability for proving left-termination of GLP's.

In Section 4.1 we claim that in order to prove left-termination of PLANNING, it is sufficient to prove acceptability of the 'part' of PLANNING called TRAS and acyclicity of the rest of the program. Let us explain how we arrive to this conclusion. First, PLANNING is partitioned into two parts: an upper part, say $P_1$ consisting of clauses $1), \ldots, 6)$, and a lower part, say $R$, consisting of the rest of PLANNING. This partition is such that no relation defined in $P_1$ occurs in $R$. This kind of partitioning of a program is defined by Apt, Marchiori and Palamidessi (1994) as follows.

Say that a *relation* is *defined in $P$* if it occurs in the head of at least one of its clauses, and that a *literal* is *defined in $P$* if its relation is defined in $P$.

**Definition 5.1 (Program Extension)** A program $P$ *extends a program $R$*, denoted by $P > R$, if no relation defined in $P$ occurs in $R$. □

So $P$ extends $R$ if $P$ defines new relations possibly using the relations defined already in $R$. For instance, the program $P_2$:

```
p ← q,r.
```

extends the program $P_1$:

```
q ← s.
s ←.
```

Next, we consider the program TRAS obtained from $P_1$ by deleting all the literals defined in $R$. We call this operation *difference*, defined as follows.

**Definition 5.2 (Difference of Two Programs)** The *difference of the programs $P$ and $R$*, denoted by $P \ominus R$, is the program obtained from $P$ by deleting all the clauses of $R$ and all the literals defined in $R$. □

For instance, if $P_1$ and $P_2$ are defined as above, then $P_2 \ominus P_1$ is the program `p← r`.

Finally, we prove that TRAS is acceptable and that $R$ is acyclic, and in doing that we have to take care that the two level mappings used are related by a condition, namely that for every ground instance, say $C = H \leftarrow Q_1, L, Q_2$, of a clause of $P_1$, for every literal $L$ contained in $C$ and defined in $R$, the level mapping of $L$ is not greater than the level mapping of $H$. This condition is important to ensure left-termination. For instance, consider the program $P$

1) `q(f(X)) ← p(Y), q(X).`
2) `p(f(X)) ← p(X).`

and take $P_1 = \{1\}$ and $R = \{2\}$. Then $P_1$ extends $R$, $P_1 \ominus R$ is acceptable w.r.t. the level mapping $|q(x)|_{P_1} = |x|$, $R$ is acyclic w.r.t. the level mapping $|p(x)|_R = |x|$, but $P$ is not left-terminating.

So, the steps we applied to PLANNING are summarized in the following definition of up-acceptability, that characterizes left-terminating programs.

For a level mapping $|\ |$ and a program $R$, the *restriction of* $|\ |$ *to* $R$, denoted $|\ |_{|R}$, is the level mapping for $R$ defined by $|A|_{|R} = |A|$.

**Definition 5.3 (Up-Acceptability)** Let $|\ |$ be a level mapping for $P$. Let $R$ be s.t. $P = P_1 \cup R$ for some $P_1$, and let $I$ be an interpretation of $P \ominus R$. $P$ is *up-acceptable w.r.t.* $|\ |$, $R$ *and* $I$ if the following conditions hold:

1. $P_1$ extends $R$;

2. $P \ominus R$ is acceptable w.r.t. $|\ |_{|P \ominus R}$ and $I$;

3. $R$ is acyclic w.r.t. $|\ |_{|R}$;

4. for every ground instance $H \leftarrow L_1, \ldots, L_n$ of a clause of $P_1$, for every $1 \leq i \leq n$,

   - if $L_i$ is defined in $R$ and is not a constraint, and
   - if $I \models L_{i1}, \ldots, L_{ik}$, where $L_{i1}, \ldots, L_{ik}$ are those literals among $L_1, \ldots, L_i$ whose relations occur in $P \ominus R$,

   then $|H| \geq |L_i|$.

A program is *up-acceptable* if there exist $|\ |$, $R$ and $I$ s.t. $P$ is up-acceptable w.r.t. $|\ |$, $R$, $I$.
□

Observe that by taking for $R$ the empty set of clauses, we obtain the original definition of acceptability. Next, we introduce the notion of *up-bounded query*.

**Definition 5.4 (Up-bounded Query)** Let $P$ be up-acceptable w.r.t. $|\ |$, $R$ and $I$, and let $Q = L_1, \ldots, L_n$. $Q$ is up-bounded if for every $1 \leq i \leq n$ the set

$$|Q|_i^{up,I} = \{|L_i'| \mid L_1', \ldots, L_n' \text{ is a ground instance of } Q \text{ and } I \models L_{k_1}' \wedge \ldots \wedge L_{k_l}'\}$$

is finite, where $L_{k_1}', \ldots, L_{k_l}'$ are the literals of $L_1', \ldots, L_{i-1}'$ whose relations occur in $P \ominus R$.
□

In order to show that all `ldcnf`-derivations of an up-bounded query are finite: we shall prove that a `ldcnf`-derivation of an up-bounded query contains only up-bounded queries; and we shall associate with each derivation of the query a descending chain in the well-founded set of pairs of multisets of natural numbers, with the lexicographic order. Recall that a *multiset* (see e.g., Deshowitz, 1987) is a unordered collection in which the number of occurrences of each element is counted. Formally, a multiset of natural numbers is a function from the set $(N, <)$ of natural numbers to itself, giving the multiplicity of each natural number. Then, the ordering $<_{mul}$ on multisets is defined as the transitive closure of the replacement of a natural number with any finite number (possibly zero) of natural numbers that are smaller under $<$. Since $<$ is well-founded, the induced ordering $<_{mul}$ is also well-founded. For simplicity we shall omit in the sequel the subscript $mul$ from $<_{mul}$.

With an up-bounded query $Q$, we associate a pair $\pi(Q)_{up,I} = (||Q||_{up,I,P_1}, ||Q||_{up,I,R})$ of multisets, where for a program $P$ and an interpretation $I$

$$||Q||_{up,I,P} = bag(max|Q|_{k_1}^{up,I}, \ldots, max|Q|_{k_m}^{up,I}),$$

where $L_{k_1}, \ldots, L_{k_m}$ are those literals of $Q$ whose relations occur in $P \ominus R$, and $max|Q|_i^{up,I}$ is the maximum of $|Q|_i^{up,I}$ (which is by convention 0 if $|Q|_i^{up,I}$ is the empty set).

Recall that the lexicographic order $\prec$ (on pairs of multisets) is defined by $(X, Y) \prec (Z, W)$ iff either $X < Z$, or $X = Z$ and $Y < W$.

Then we can prove the following result.

**Theorem 5.5** *Suppose that $P$ is up-acceptable w.r.t. $||$, $R$ and $I$. Let $Q$ be an up-bounded query. Then every `ldcnf`-derivation for $Q$ in $P$ contains only up-bounded queries and is finite.*

*Proof.* Let $\xi = Q_1, \ldots, Q_n, \ldots$ be a `ldcnf`-derivation for $Q$ in $P$. We prove by induction on $n$ that $Q_n$ is up-bounded, and that if it is the resolvent of a query $Q_{n-1}$ by the selection of a literal which is not a constraint, then $\pi(Q_n)_{up,I} < \pi(Q_{n-1})_{up,I}$.

For the base case $n = 1$, we have that $Q_1$ is up-bounded by assumption. Now consider $n > 1$, and suppose that the result holds for $n - 1$. Thus, $Q_{n-1}$ is up-bounded. Suppose that the resolvent of $Q_{n-1}$ is defined and that the selected literal, say $L$, is not a constraint. It follows from the fact that $Q_{n-1}$ is up-bounded and from the definition of up-acceptability (here also condition 4 is used) that $Q_n$ is up-bounded. Next, we show that $\pi(Q_n)_{up,I}$ is smaller than $\pi(Q_{n-1})_{up,I}$ in the lexicographic order. If the relation symbol of $L$ occurs in $P \ominus R$ then the first component of $\pi(Q_n)_{up,I}$ becomes smaller because of condition 2. Otherwise, if the relation symbol of $L$ occurs in $R$ then the first component of $\pi(Q_n)_{up,I}$ does not increase because of condition 1, while the second one becomes smaller because of condition 3. The conclusion follows from the fact that the lexicographic ordering is well-founded, and from the fact that, in a derivation a constraint can be consecutively selected only a finite number of times. □

**Example 5.6** (planning is Up-Acceptable) Call R-blocksworld the program obtained from blocksworld by deleting the clauses 6) and 7). We prove that planning is up-acceptable w.r.t. $||$, R-blocksworld, and $I$ defined as in the examples of Sections 3.1 and 4.1. PLANNING⊖R-BLOCKSWORLD is (not incidentally) the program TRAS. The proof of up-acceptability proceeds as follows.

1. PLANNING extends R-BLOCKSWORLD.

2. It is proven in Section 4.1 that TRAS is acceptable.

3. It is proven in Section 3.1 that R-BLOCKSWORLD acyclic.

4. Consider a ground instance
   $transform(c, s, p) \leftarrow state(s0), legals(s0, c), trans(c, s, [s0], p).$
   of 1), and suppose that $I \models state(s0)$. Then
   $|transform(c, s, p)| = tot + 3 * (|c| + 1) + 6 \geq 3 * |c| + 2 = |legals(s0, c)|.$
   Consider a ground instance
   $trans(c, s, v, [\,]) \leftarrow legals(s, c).$
   of 1). Then
   $|trans(c, s, v, [\,])| = tot - card(el(v) \cap S) + 3 * (|c| + 1) + 5 + |v| \geq 3 * |c| + 2.$

   □

The following corollary establishes the equivalence of the notions of acceptability and up-acceptability. It follows directly from Theorem 5.5 and Theorem 4.6.

**Corollary 5.7** *A general logic program is up-acceptable if and only if it is acceptable.*

## 6. Weak Up-Acceptability

Because in some cases up-acceptability does not help to simplify the proof of termination, in this section we generalize this notion and introduce weak up-acceptability. We start with an example of a program that cannot be split into two non-empty programs satisfying up-acceptability. Next, we introduce weak up-acceptability and establish analogous results as for up-acceptability. Finally, we apply weak up-acceptability for simplifying the proof of left-termination of our example program.

### 6.1 An Example: Hamiltonian Path

A Hamiltonian path of a graph is an acyclic path containing all the nodes of the graph. The following program HAMILTONIAN defines hamiltonian paths: it consists of the following clauses

```
1) ham(G,P) ←
       path(N1,N2,G,P),
       cov(P,G).
2) cov(P,G) ←
       ¬ notcov(P,G).
3) notcov(P,G) ←
```

```
        node(X,G), ¬ member(X,P).
4) node(X,G) ←
        member([X,Y],G).
5) node(X,G) ←
        member([Y,X],G).
```

augmented with the program ACYPATH defining acyclic paths:

```
p1) path(N1,N2,G,P) ←
        path1(N1,[N2],G,P).
p2) path1(N1,[N1|P1],G,[N1|P1]) ←.
p3) path1(N1,[X1|P1],G,P) ←
        member([Y1,X1],G),
        ¬ member(Y1,[X1|P1]),
        path1(N1,[Y1,X1|P1],G,P).
p4) member(X,[X|Y]) ←.
p5) member(X,[Y|Z]) ←
        member(X,Z).
```

A graph is represented by means of a list of edges. For graphs consisting only of one node, we adopt the convention that they are represented by the list $[[a, \perp]]$, where $\perp$ is a special new symbol. In the clause p1) $path$ describes acyclic paths of a graph, and $path(n1, n2, g, p)$ calls the query $path1(n1, [n2], g, p)$. The second argument of $path1$ is used to construct incrementally an acyclic path connecting $n1$ with $n2$: using clause p3), the partial path $[x|p1]$ is transformed into $[y, x|p1]$ if there is an edge $[y, x]$ in the graph $g$ such that $y$ is not already present in $[x|p1]$. The construction terminates if $y$ is equal to $n1$, because of clause p2). Thus the relation $path1$ is defined inductively by the clauses p2) and p3), using the familiar relation $member$, specified by the clauses p4) and p5). Notice that, it follows from p2) that if $n1$ and $n2$ are equal, then $[n1]$ is assumed to be an acyclic path from $n1$ to $n2$, for any $g$.

The relation $ham(g, p)$ is specified in terms of $path$ and $cov$: it is true if $p$ is an acyclic path of $g$ that covers all its nodes. The relation $cov$ is defined as the negation of $notcov$, where $notcov(p, g)$ is true if there is a node of $g$ which does not occur in $p$. Finally, the relation $node$ is defined in terms of $member$ in the expected way. For instance, `ham([[a,b],[b,c],[a,a],[c,b]], [a,b,c])` holds, corresponding to the path drawn in bold in the graph of Figure 3.

The program HAMILTONIAN is not terminating, because ACYPATH is not. However, HAMILTONIAN is left-terminating. In order to prove this result using acceptability (Definition 4.2), we need to find a model of HAMILTONIAN that is also a model of the completion $comp(\{3), 4), 5), p4), p5)\})$ of the program consisting of the clauses 3), 4), 5), p4), p5). This is not very difficult, however it is not needed, as we shall see in the follow. Note also that the notion of up-acceptability does not help to prove left-termination using less semantic information. Nevertheless, we can split HAMILTONIAN in two subprograms: $P_2$ consisting of ACYPATH plus clause 1), and $P_1$ consisting of the remaining clauses 2) − 5). Note that $P_2$ 'almost' extends $P_1$, because $P_1$ contains some literals (those with relation $\{member\}$) defined in $P_2$. Since the subprogram $5p), 6p)$ defining these literals is extended by both $P_1$ and by $P_2 \setminus \{5p), 6p)\}$, it follows that left-termination of $\{5p), 6p)\}$ does not depend on the termination behaviour of the rest of HAMILTONIAN. So, for proving left-termination of HAMILTONIAN it is sufficient to show that $P_2 \ominus P_1$ is acceptable, that $P_1$ is acyclic, and that the corresponding level mappings satisfy the condition in Definition 5.3. Thus, we need only to find a model of $P_2 \ominus P_1$ that is also a model of $comp(\{p4), p5)\})$). □

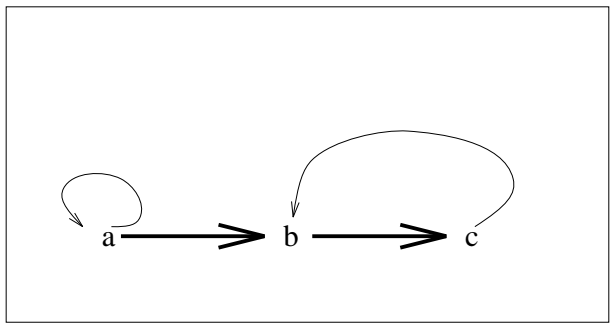

Figure 3: The Hamiltonian path of $[[a, b], [b, c], [a, a], [c, b]]$

## 6.2 Weak Up-Acceptability

Formally, we modify up-acceptability by considering a more general way of partitioning the program, specified using the following notion of weak extension. Recall that for a set $S$ of relations, $P_{|S}$ denotes the clauses of $P$ that define the relations from $S$.

**Definition 6.1 (Program Weak Extension)** A program $P$ *weakly extends a program* $R$, denoted by $P >_w R$, if for some set $S$ of relations we have that:

- $P = P_1 \cup P_{|S}$, and $P_1$ extends $P_{|S}$;

- $R$ extends $P_{|S}$; and

- $P \ominus P_{|S}$ extends $R \ominus P_{|S}$. □

Note that only the relations of $S$ which are defined in $P$ play a role in the above definition. Definition 5.1 is a particular case of the above definition, obtained by considering $P_{|S}$ to be equal to $\emptyset$ (which includes the case that $S = \emptyset$).

**Example 6.2** The program

```
p(X) ← q(X), r(X).
r(f(X)) ← r(X).
```

weakly extends the program

```
q(X) ← s(X), r(X).
s(X) ←.
```

This can be seen by taking $S = \{r\}$. Then $P_1$ is `p(X) ← q(X), r(X).`, $P_{|S}$ is `r(f(X)) ← r(X).`, $P_1$ and $R$ both extend $P_{|S}$. Moreover, $P \ominus P_{|S}$ is `p(X) ← q(X).` and $R \ominus P_{|S}$ is

```
q(X) ← s(X).
s(X) ←.
```

Finally, it is easy to check that $P \ominus P_{|S}$ extends $R \ominus P_{|S}$.  □

Thus the notion of weak up-acceptability is obtained from Definition 5.3 by replacing in condition 1 'extends' by 'weakly extends'.

**Definition 6.3 (Weak Up-Acceptability)** Let $|\ |$ be a level mapping for $P$. Let $R$ be a set of clauses s.t. $P = P_1 \cup R$ for some $P_1$, and let $I$ be an interpretation of $P \ominus R$. $P$ is *weakly up-acceptable w.r.t.* $|\ |$, $R$ *and* $I$ if the following conditions hold:

1. $P_1$ weakly extends $R$;

2. $P \ominus R$ is acceptable w.r.t. $|\ |_{|P \ominus R}$ and $I$;

3. $R$ is acyclic w.r.t. $|\ |_{|R}$;

4. for every ground instance $H \leftarrow L_1, \ldots, L_n$ of a clause of $P_1$, for every $1 \leq i \leq n$,

   - if $L_i$ is defined in $R$ and is not a constraint, and
   - if $I \models L_{i1}, \ldots, L_{ik}$, where $L_{i1}, \ldots, L_{ik}$ are those literals among $L_1, \ldots, L_i$ whose relations occur in $P \ominus R$,

   then $|H| \geq |L_i|$.  □

In order to prove the analog to Theorem 5.5, we need to use triples of finite multisets, instead of pairs, with the lexicographic ordering $\prec$: $(X_1, X_2, X_3) \prec (Y_1, Y_2, Y_3)$ iff either $(X_1, X_2) \prec (Y_1, Y_2)$ (by abuse of notation we use $\prec$ also to denote the lexicographic ordering on pairs of multisets), or $X_1 = Y_1$ and and $X_2 = Y_2$ and $X_3 < Y_3$. We consider the triple:

$$\tau(Q)_{up,I} = (|[Q]|_{up,I,P \ominus P_{|S}}, |[Q]|_{up,I,R \ominus P_{|S}}, |[Q]|_{up,I,P_{|S}}).$$

**Theorem 6.4** *Suppose that $P$ is weakly up-acceptable w.r.t.* $|\ |$, $R$ *and* $I$. *Let* $Q$ *be an up-bounded query. Then every* `ldcnf`*-derivation for* $Q$ *in* $P$ *contains only up-bounded queries and is finite.*

*Proof.* Let $S$ be the set of relations used to prove that $P$ is weakly up-acceptable w.r.t. $|\ |$, $R$ and $I$. The proof is similar to the one of Theorem 5.5, except that we consider $\tau(Q)_{up,I}$ instead of $\pi(Q)_{up,I}$, and we show that $\tau(Q_n)_{up,I}$ is smaller than $\tau(Q_{n-1})_{up,I}$ in the lexicographic order as follows. If the relation symbol of $L$ occur in $P \ominus R$ but not in $S$, then the first component of $\tau(Q_n)_{up,I}$ becomes smaller because of condition 2. Otherwise, if the relation symbol of $L$ occur in $R$ then the first component of $\tau(Q_n)_{up,I}$ does not increase because of condition 1, while the second one becomes smaller because of condition 3. Finally, if the relation symbol of $L$ occur in $S$, then the first and second components of $\tau(Q_n)_{up,I}$ do not increase, because of condition 1, while the third one becomes smaller because of condition 2. □

**Example 6.5** (HAMILTONIAN is Weakly Up-Acceptable) We prove that HAMILTONIAN is weakly up-acceptable. Consider as upper part the program $P_2$ consisting of ACYPATH augmented with clause 1), and as lower part the program $P_1$:

```
2) cov(P,G) ←
       ¬ notcov(P,G).
3) notcov(P,G) ←
       node(X,G), ¬ member(X,P).
4) node(X,G) ←
       member([X,Y],G).
5) node(X,G) ←
       member([Y,X],G).
```

Take $\{member\}$ as set $S$ of relations.

1. $P_2$ weakly extends $P_1$.

2. The program $P_2 \ominus P_1$, consisting of

```
1) ham(G,P) ←
       path(N1,N2,G,P).
p1) path(N1,N2,G,P) ←
       path1(N1,[N2],G,P).
p2) path1(N1,[N1|P1],G,[N1|P1]) ←.
p3) path1(N1,[X1|P1],G,P) ←
       member([Y1,X1],G),
       ¬ member(Y1,[X1|P1]),
       path1(N1,[Y1,X1|P1],G,P).
p4) member(X,[X|Y]) ←.
p5) member(X,[Y|Z]) ←
       member(X,Z).
```

is acceptable w.r.t. the following level mapping:

$|member(s,t)| = |t|;$

$|path1(n1,p1,g,p)| = |p1| + |g| + 2(|g| - |p1 \cap g|) + 1;$

$|path(n1,n2,g,p)| = 3|g| + 3;$

$|ham(g,p)| = 3|g| + 4,$

and the interpretation $I = I_{ham} \cup I_{path} \cup I_{path1} \cup I_{member}$, with:

$I_{ham} = [ham(G, P)]$,

$I_{path} = \{path(n1, n2, g, p) \mid |g| + 1 \geq |p|\}$,

$I_{path1} = \{path1(n1, p1, g, p) \mid |p1| - |p1 \cap g| \geq |p| - |p \cap g|\}$,

$I_{member} = \{member(s, t) \mid t \text{ list s.t. } s \in set(t)\}$,

where for two lists $p$ and $g$, $p \cap g$ denotes the list containing as elements those $x$ which are elements of $p$ for which there exists a $y$ s.t. $[x, y]$ is an element of $g$.

We prove that $I$ is a model of $P_2$.

- Consider a ground instance of the clause p1) and suppose that

$$I \models path1(n1, [n2], g, p).$$

  Note that $|[n2]| - |[n2] \cap g| \leq 1$. So $|p| - |p \cap g| \leq 1$. But $|p \cap g| \leq |g|$. Then $|p| \leq |g| + 1$, hence $I \models path(n1, n2, g, p)$.

- Consider a ground instance of the clause p3) and suppose that

$$I \models member([y1, x1], g), \neg member(y1, [x1|p1]), path1(n1, [y1, x1|p1], g, p).$$

  Thus $|[y1, x1|p1]| - |[y1, x1|p1] \cap g| \geq |p| - |p \cap g|$, where $y1 \notin [x1|p1]$ and $[y1, x1] \in g$. Therefore $|[y1, x1|p1] \cap g| = 1 + |[x1|p1] \cap g|$. So $|[y1, x1|p1]| - |[y1, x1|p1] \cap g| = |[x1|p1]| - |[x1|p1] \cap g|$. Then $|[x1|p1]| - |[x1|p1] \cap g| \geq |p| - |p \cap g|$. Hence $I \models path1(n1, [x1|p1], g, p)$.

- The proof for the other clauses is analogous.

Now, $Neg_{P_2} = \{member\}$ and $P_2^- = \{(f), (g)\}$. It is routine to check that $I_{|\{member\}}$ is a model of $comp(P_2^-)$.

3. $P_1$ is acyclic w.r.t. the level mapping:

$|cov(p, g)| = |p| + |g| + 3$;

$|notcov(p, g)| = |p| + |g| + 2$;

$|node(s, t)| = |t| + 1$;

$|member(s, t)| = |t|$.

4. Consider a ground instance

$$ham(g, p) \leftarrow path(n1, n2, g, p), cov(p, g).$$

of 1) and suppose that $I \models path(n1, n2, g, p)$. So $|g| + 1 \geq |p|$. Hence $|ham(g, p)| = 3|g| + 4 \geq |p| + |g| + 3 = |cov(p, g)|$. $\qquad \square$

## 7. Low-Acceptability

In the previous two sections, we have integrated the notions of acyclicity and acceptability, by means of a partition of the program into an upper and a lower part. We introduced the notion of up- and weak up-acceptability, where the upper part of the program is proven to be acceptable and the lower part acyclic. In order to treat also the converse case, i.e., the upper part being acyclic and the lower part acceptable, we introduce now the notion of low-acceptability. We follow the structure of the previous sections: first, a motivating example is presented. Next, we define the notion of low-acceptability and prove some results. Finally, we apply this notion to the program of our example.

### 7.1 An Example: Graph Specialization

Graph structures are used in AI for many applications, such as the representation of relations, situations or problems (see e.g., Bratko, 1986). Two typical operations on graphs are *find a path between two given nodes*, and *find a subgraph with some specified properties*. The program SPECIALIZE below uses both these operations to solve the following problem. Given two nodes $n_1, n_2$ in a graph $g$, find a node $n$ that does not belong to any acyclic path in $g$ from $n_1$ to $n_2$. The program SPECIALIZE consists of the clauses:

```
1) spec(N1,N2,N,G) ←
      ¬ unspec(N1,N2,N,G).
2) unspec(N1,N2,N,G) ←
      path(N1,N2,G,P),
      member(N,P).
```

augmented with the program ACYPATH of the previous section. The relation *spec* is specified as the negation of *unspec*, where $unspec(n1, n2, n, g)$ is true if there is an acyclic path of the graph $g$ connecting the nodes $n1$ and $n2$ and containing $n$. For instance, `spec(a,b,c,[[a,b],[b,c],[a,a],[c,b]])` holds (Figure 4).

Observe that SPECIALIZE is not terminating: for instance, the query `path1(a,[b,c],d,e)` has an infinite derivation obtained by choosing as input clause (a variant of) the clause $p3$) and by selecting always its rightmost literal. However SPECIALIZE is left-terminating. In order to prove this result using acceptability (Definition 4.2), we need to find a model of SPECIALIZE that is also a model of *comp*(SPECIALIZE), which is rather difficult. Note also that the notions of weak up- and up-acceptability do not help to simplify the proof. However, we can split SPECIALIZE in two subprograms: $P_2$ consisting of the clause 1) and $P_1$ consisting of the rest of the program. Note that $P_2$ extends $P_1$. Therefore, in order to show that SPECIALIZE is left-terminating, it is sufficient to prove that $P_2 \ominus P_1$ is acyclic, that $P_1$ is acceptable, and that the corresponding level mappings are suitably related.

### 7.2 Low-Acceptability

Formally, we introduce the following notion of low-acceptability.

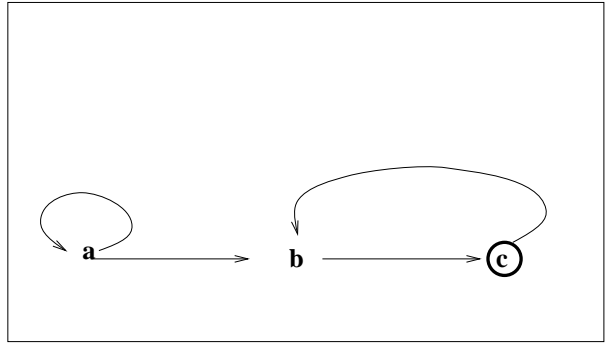

Figure 4: $spec(a, b, c, [[a, b], [b, c], [a, a], [c, b]])$ holds

**Definition 7.1 (Low-Acceptability)** Let $|\,|$ be a level mapping for $P$. Let $R$ be a set of clauses s.t. $P = P_1 \cup R$ for some $P_1$, and let $I$ be an interpretation of $R$. $P$ is *low-acceptable w.r.t.* $|\,|$, *R and I* if the following conditions hold:

1. $P_1$ extends $R$;

2. $P \ominus R$ is acyclic w.r.t. $|\,|_{|P \ominus R}$;

3. $R$ is acceptable w.r.t. $|\,|_{|R}$ and $I$;

4. for every ground instance $H \leftarrow L_1, \ldots, L_n$ of a clause of $P_1$, for every $1 \leq i \leq n$, if $L_i$ is defined in $R$ and is not a constraint, then $|H| \geq |L_i|$.

A program is *low-acceptable* if there exist $|\,|$, $R$ and $I$ s.t. $P$ is low-acceptable w.r.t. $|\,|$, $R$ and $I$. □

The notion of *low-boundedness* is defined as in the previous section, by replacing $|Q|_i^{up,I}$ with

$$|Q|_i^{low,I} = \{|L_i'| \mid L_1', \ldots, L_n' \text{ is a ground instance of } Q \text{ and } I \models L_{k_1}' \wedge \ldots \wedge L_{k_l}'\},$$

where $L_{k_1}', \ldots, L_{k_l}'$ are the literals of $L_1', \ldots, L_{i-1}'$ whose relations occur in $R$.

To prove the analogue of Theorem 5.5 for low-bounded queries, we associate with a low-bounded query $Q$ a pair $\pi(Q)_{low,I} = (||Q||_{low,I,P_1}, ||Q||_{low,I,R})$ of multisets, with for a program $P$ and an interpretation $I$

$$||Q||_{low,I,P} = bag(max|Q|_{k_1}^{low,I}, \ldots, max|Q|_{k_m}^{low,I}),$$

where $L_{k_1}, \ldots, L_{k_m}$ are the literals of $Q$ whose relations occur in $P$.

**Theorem 7.2** *Suppose that $P$ is low-acceptable w.r.t. $|\,|$, $R$ and $I$. Let $Q$ be a low-bounded query. Then every* `ldcnf`*-derivation for $Q$ in $P$ contains only low-bounded queries and is finite.*

*Proof.* The proof is similar to that of Theorem 5.5, where one replaces $\pi(Q)_{up,I}$ with $\pi(Q)_{low,I}$. □

The following result is a direct consequence of Theorems 7.2 and 4.6.

**Corollary 7.3** *A general logic program is low-acceptable if and only if it is acceptable.*

**Example 7.4** (SPECIALIZE is Low-Acceptable) We show that the program SPECIALIZE is low-acceptable. Consider the program SPEC1=SPECIALIZE\\{1)}. Then the proof proceeds as follows.

1. The program {1)} extends SPEC1.

2. The program {1)}⊖SPEC1 is acyclic w.r.t. the level mapping
   $$|spec(n1, n2, n, g)| = 3|g| + 5.$$

3. The program SPEC1 is acceptable w.r.t. | | and the interpretation $I$, with | | defined as in Example 6.5 for atoms with relation $member$, $path1$, $path$, and $|unspec(n1, n2, n, g)| = 3|g| + 4$; and with $I = I_{unspec} \cup I_{path} \cup I_{path1} \cup I_{member}$, s.t.:
   $$I_{unspec} = [unspec(N1, N2, N, G)],$$
   and $I_{path}$, $I_{path1}$, and $I_{member}$ are as before (Example 6.5).

4. Consider a ground instance
   $$spec(n1, n2, n, g) \leftarrow \neg unspec(n1, n2, n, g)$$
   of 1). Then
   $$|spec(n1, n2, n, g)| = 3|g| + 5 \geq 3|g| + 4 = |unspec(n1, n2, n, g)|.$$

Consider the query $Q = \texttt{spec(a,b,X,[[a,b],[b,c],[a,a]])}$. Because $Q$ is low-bounded, it has a finite `ldcnf`-tree, with answer $X \neq a, X \neq b$. Notice that by using negation as failure $Q$ flounders. □

## 8. A Methodology for Proving Left-Termination

Definitions 5.3, 6.3 and 7.1 provide a method for proving left-termination of a GLP, which is summarized in Definition 8.1 below. In this section, we first discuss advantages and drawbacks of this method. Next, we introduce a methodology for proving left-termination of GLP's that incorporates the notions we have introduced in the previous sections. Finally, we give an example in order to illustrate the methodology.

**Definition 8.1** (A Method for Proving Left-Termination)

1. Find a maximal set $R$ of clauses of $P$ s.t. $R$ forms an acyclic program and $P = P_1 \cup R$ is s.t. either $R$ extends $P_1$ or vice versa.

2. If $R$ extends $P_1$ then:

(a) Prove that $P \ominus R$ is acceptable w.r.t. a level mapping, say $|\ |_{P \ominus R}$, and an interpretation.

(b) Use $|\ |_{P \ominus R}$ to define a level mapping $|\ |_R$ for $R$ s.t. $R$ is acyclic w.r.t. $|\ |_R$, and s.t. for every ground instance $H \leftarrow L_1, \ldots, L_n$ of a clause of $R$, for every $1 \leq i \leq n$: if $L_i$ is defined in $P_1$ then $|H|_R \geq |L_i|_{P \ominus R}$ holds.

3. If $P_1$ extends $R$ then:

(a) Prove that $R$ is acyclic w.r.t. a level mapping, say $|\ |_R$.

(b) Use $|\ |_R$ to define a level mapping $|\ |_{P \ominus R}$ for $P \ominus R$ s.t. $P \ominus R$ is acceptable w.r.t. $|\ |_{P \ominus R}$ and an interpretation $I$, and s.t. for every ground instance $H \leftarrow L_1, \ldots, L_n$ of a clause of $P_1$, for every $1 \leq i \leq n$: if $L_i$ is defined in $R$ and if those literals among $L_1, \ldots, L_i$ whose relations occur in $P \ominus R$, say $L_{i1}, \ldots, L_{ik}$, are s.t. $I \models L_{i1}, \ldots, L_{ik}$, then $|H|_{P \ominus R} \geq |L_i|_R$ holds. $\square$

An advantage if this method is that it partly overcomes a drawback of the original method of Apt and Pedreschi to prove left-termination, where one has to find a specialized model of the *entire* program. Unfortunately, our method is not always applicable. This happens because in point 2. we use $P \ominus R$, thus discarding the literals of $R$ occurring in $P_1$. These literals could be relevant for the left-termination behaviour of $P_1$. For instance, in the program

```
p ← q, p.
q ← s.
```

if we take $P_1$ and $R$ to be the first and second clause, respectively, then $P_1$ extends $R$, but $P_1 \ominus R$ is p← p, a clearly non-acceptable program. This problem can be overcome by considering also some semantic information about $R$, which leads to the following alternative definition of up-acceptability.

**Definition 8.2 (New Up-Acceptability)** Let $|\ |$ be a level mapping for $P$. Let $R$ be s.t. $P = P_1 \cup R$ for some $P_1$, let $I_R$ be a specialized model of $R$, and let $I_{P_1}$ be a specialized model of $P \ominus R$. $P$ is *new up-acceptable w.r.t.* $|\ |$, $R$, $I_R$ and $I_{P_1}$ if the following conditions hold:

1. $P_1$ extends $R$;

2. for all ground instances $H \leftarrow L_1, \ldots, L_n$ of clauses of $P_1$, for every $1 \leq i \leq \overline{n}$, with

$$\overline{n} = min(\{n\} \cup \{j \in [1, n] \mid I_R \cup I_{P_1} \not\models L_j\}),$$

- if $L_i$ is defined in $P \ominus R$ then $|H| > |L_i|$,
- if $L_i$ is defined in $R$ then $|H| \geq |L_i|$.

3. $R$ is acyclic w.r.t. $|\ |$. $\square$

202

One can check that the results we proved for up-acceptability hold as well for the above definition. In particular, the notion of new up-acceptability is equivalent to the one of acceptability. Note that here we have to find some semantic information on both the 'upper' and the 'lower' part of the program; however, information on the 'lower' part is used only on the 'upper' part of the program. Therefore, also in this case, less semantic information is needed than with the original definition of acceptability by Apt and Pedreschi. Let us illustrate the application of new up-acceptability in the following toy example.

**Example 8.3** Consider again the program

```
1) p ← q, p.
2) q ← s.
```

We prove that it is new up-acceptable.

1. The program $\{1)\}$ extends $\{2)\}$;

2. Consider the level mapping

   $|p| = 1, |q| = 1, |s| = 0$,

   and the interpretations

   $I_{\{1\}} = \{p\}, I_{\{2\}} = \emptyset$.

   Then $I_{\{1\}}$ and $I_{\{2\}}$ are specialized models of $\{1)\}$ and of $\{2)\}$, respectively. We have that $I_{\{1\}} \cup I_{\{2\}} \not\models q$ and $|p| = |q|$.

3. From $|q| = 1 > 0 = |s|$ it follows that $\{2)\}$ is acyclic w.r.t. $|\ |$. □

Observe that Definition 8.2 is still not applicable in some cases, for instance to the program

```
1) p ← q, ¬ p.
2) q ← s.
```

because the program $\{1)\} \ominus \{2)\}$ has no specialized model.

Another drawback of our method is its lack of incrementality. Nevertheless, we can define an incremental, bottom-up method, where the decomposition step is applied iteratively to the subprograms until the partition of a subprogram becomes trivial. This is possible because of the equivalence of up-/weak up-/ low-acceptability and acceptability. These observations are incorporated in the following definition. Recall that $\mathcal{B}_P$ denotes the Herbrand base of $P$.

**Definition 8.4 (An Incremental Method)**

- Split $P$ into $n \geq 1$ parts, say $P_1, \ldots, P_n$ s.t. for every $i \in [1, n-1]$:

  - $P_{i+1}$ (weakly) extends $P_i$;
  - either $P_i$ or $P_{i+1}$ is acyclic.

- Define incrementally the level mapping $|\,|_{P_1 \cup \ldots \cup P_n} = |\,|_{P_1} \cup \ldots \cup |\,|_{P_n}$ and the interpretation $I_{P_1 \cup \ldots \cup P_n} = I_{P_1} \cup \ldots \cup I_{P_n}$ as follows.

  1. (base) If $P_1$ is acyclic then find the corresponding level mapping $|\,|_{P_1}$; otherwise prove that $P_1$ is acceptable w.r.t. a level mapping $|\,|_{P_1}$ and an interpretation $I_{P_1}$.

  2. (induction) Suppose that $|\,|_{P_k}$ is defined for every $1 \leq k \leq i$, and suppose that $I_{P_k}$ is defined for every $1 \leq k < i$ if $P_i$ is acyclic, and for every $1 \leq k \leq i$ if $P_i$ is acceptable, with $1 \leq i < n$. Then,

     (a) If $P_{i+1}$ is acyclic then use $|\,|_{P_i}$ to define a level mapping $|\,|_{P_{i+1}}$ for $P_{i+1} \ominus P_i$ s.t. $P_{i+1} \ominus P_i$ is acyclic w.r.t. $|\,|_{P_{i+1}}$, and s.t. for all ground instances $H \leftarrow L_1, \ldots, L_m$ of clauses of $P_{i+1}$, for every $1 \leq j \leq m$, if $L_j$ is defined in $P_i$ then $|H|_{P_{i+1}} \geq |L_j|_{P_i}$.

     (b) If $P_i$ is acyclic then use $|\,|_{P_i}$ to define a level mapping $|\,|_{P_{i+1}}$ for $P_{i+1} \ominus P_i$ s.t.:

        i. A. either $P_{i+1} \ominus P_i$ is acceptable w.r.t. a specialized model $I_{P_{i+1}}$ and $|\,|_{P_{i+1}}$; in this case set $I_{P_i}$ to be $\mathcal{B}_{P_i}$;

           B. or find a specialized model $I_{P_i}$ of $P_i \ominus P_{i-1}$, and a specialized model $I_{P_{i+1}}$ of $P_{i+1} \ominus P_i$ s.t. for all ground instances $H \leftarrow L_1, \ldots, L_m$ of clauses of $P_{i+1}$ and for every $1 \leq k \leq \overline{m}$ if $L_k$ is defined in $P_{i+1}$ then $|H|_{P_{i+1}} > |L_k|_{P_{i+1}}$.

        ii. For all ground instances $H \leftarrow L_1, \ldots, L_m$ of clauses of $P_{i+1}$ and for every $1 \leq k \leq \overline{m}$ if $L_k$ is defined in $P_i$ then $|H|_{P_{i+1}} \geq |L_k|_{P_i}$.

        Above, $\overline{m} = min(\{m\} \cup \{j \in [1, m] \mid I_{P_1 \cup \ldots \cup P_{i+1}} \not\models L_j\})$.

$\square$

We prove that this method is correct, i.e., that $P$ is left-terminating if the above method is applicable. To deal with non-ground queries, we use the original notion of boundedness by Apt and Pedreschi, this time w.r.t. the interpretation resulting from the method.

**Definition 8.5 (Bounded Query)** Suppose that the partition $P_1, \ldots, P_n$ of $P$, $|\,|_{P_1 \cup \ldots \cup P_n}$ and $I_{P_1 \cup \ldots \cup P_n}$ are obtained using the method of Definition 8.4. Let $Q = L_1, \ldots, L_m$. Then $Q$ is *bounded (w.r.t. $|\,|$ and $I_{P_1 \cup \ldots \cup P_n}$)* if for every $1 \leq i \leq m$, the set

$$|Q|_i^{I_{P_1 \cup \ldots \cup P_n}} = \{|L_i'| \mid L_1', \ldots, L_n' \text{ is a ground instance of } Q \text{ and}$$
$$I_{P_1 \cup \ldots \cup P_n} \models L_1' \wedge \ldots \wedge L_{i-1}'\}$$

is finite. $\square$

**Theorem 8.6** *Suppose that the partition $P_1, \ldots, P_n$, $|\,|_{P_1 \cup \ldots \cup P_n}$ and $I_{P_1 \cup \ldots \cup P_n}$ are obtained using the method of Definition 8.4. Let $Q$ be a bounded query w.r.t. $|\,|_{P_1 \cup \ldots \cup P_n}$ and $I_{P_1 \cup \ldots \cup P_n}$. Then every* `ldcnf`*-derivation of $Q$ is finite and it contains only bounded queries.*

*Proof.* Recall that $I_{P_1 \cup \ldots \cup P_n} = I_{P_1} \cup \ldots \cup I_{P_n}$. For a bounded query $Q = Q_1, \ldots, Q_m$, we define the $n$-tuple $\pi(Q)_{I_{P_1 \cup \ldots \cup P_n}} = (||Q||_{I_{P_n}, P_n \ominus P_{n-1}}, \ldots, ||Q||_{I_{P_2}, P_2 \ominus P_1}, ||Q||_{I_{P_1}, P_1})$ of multisets, with for a program $P$, and an interpretation $I$, $||Q||_{I,P} = bag(max|Q|_{k_1}^I, \ldots, max|Q|_{k_m}^I)$,

where $L_{k_1}, \ldots, L_{k_m}$ are the literals of $Q$ whose relations occur in $P$. The proof is similar to the one of Theorem 5.5. □

In the following section we illustrate the application of this method.

## 8.1 An Example: Graph Reduction

In Example 7.4, a program is described which for a graph $g$ and two nodes $n1$ and $n2$, finds a node $n$ that does not belong to any acyclic path in $g$ from $n1$ to $n2$. Using this program, we define here the program REDUCE which for a non-empty graph $g$ and two nodes $n1$ and $n2$, computes the graph $g'$ obtained from $g$ by removing all the nodes that do not belong to any acyclic path in $g$ from $n1$ to $n2$, and all the arcs containing at least one of such nodes (see Figure 5).

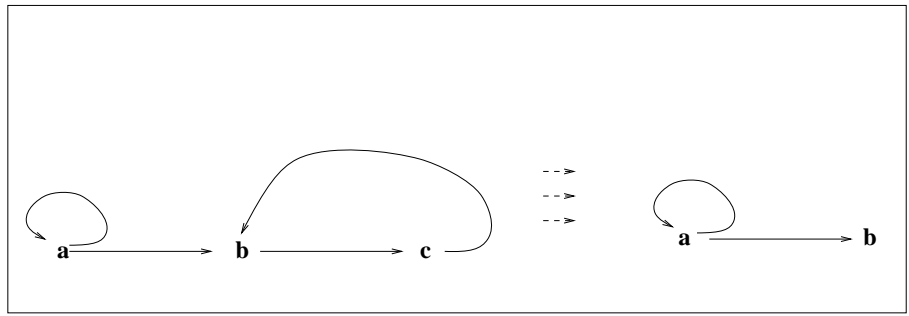

Figure 5: $rem(a, b, [[a, b], [b, c], [a, a], [c, b]], [[a, b], [a, a]])$ holds

The program REDUCE consists of the clauses:

```
1) red(N1,N2,G1,G2) ←
      ¬ unif(G1,[]),
      spec(N1,N2,N,G1),
      rem(N,G1,G),
      red(N1,N2,G,G2).
2) red(N1,N2,G,G) ←
      ¬ spec(N1,N2,N,G).
3) rem(N,[[X,Y]|G1],G2) ←
      member(N,[X,Y]),
      rem(N,G1,G2).
4) rem(N,[[X,Y]|G1],[[X,Y]|G2]) ←
      ¬ member(N,[X,Y]),
      member(N,G1),
      rem(N,G1,G2).
5) rem(N,[],[]) ←.
6) unif(G,G) ←.
```

plus the program SPECIALIZE. The relation $red(n1, n2, g, g')$ is defined by two mutually exclusive cases, corresponding to the clauses 1) and 2). Clause 1) describes the case where there is a node that does not belong to any acyclic path in $g$ from $n1$ to $n2$: first, the relation *spec* is used to find such a node; next, the node and the corresponding arcs are deleted from the graph, using the relation *rem*; finally, *red* is called recursively on the resulting graph. Clause 2) describes the final situation, where $g$ contains only nodes that belong to some of its acyclic paths from $n1$ to $n2$. The relation $rem(n, g1, g2)$ holds if the graph $g2$ is obtained from the graph $g1$ by deleting all the arcs containing the node $n$ of $g1$. It is recursively defined by the clauses 3), 4) and 5), as one would expect.

Observe that queries of the form $red(n1, n2, [\ ], g)$ fail, for every $n1, n2, g$.

We prove that REDUCE is left-terminating by using our bottom-up method. REDUCE can be partitioned in three parts:

- $P_1$ is the program SPEC1 of Example 7.4;

- $P_2$ consists of the clauses 3), 4), 5) of REDUCE plus the clauses 1), p4), p5) of SPE-CIALIZE;

- $P_3$ consists of the clauses 1), 2), and 6) of REDUCE.

It is easy to check that $P_2$ is acyclic. Moreover, $P_3$ extends $P_2$, and $P_2$ weakly extends $P_1$ w.r.t. $\{member\}$. So we can apply the bottom-up approach to construct a level mapping $|\ |_{P_1 \cup P_2 \cup P_3}$ and an interpretation $I_{P_1 \cup P_2 \cup P_3}$. The proof proceeds as follows.

- $P_1$ is acceptable w.r.t. $|\ |_{P_1}$ and $I_{P_1}$ given in Example 7.4.

- $P_2 \ominus P_1$ is acyclic w.r.t. $|\ |_{P_2}$ defined as in Example 7.4 for *spec* and *member*, and s.t. $|rem(n, g1, g2)|_{P_2} = |g1| + 2$.

  Moreover, clause 1) of SPECIALIZE satisfies the condition relating the two level mappings.

- In order to define $|\ |_{P_3}$, $I_{P_3}$ and $I_{P_2}$, we apply point i.B. Consider the level mapping
  $|red(n1, n2, g1, g2)|_{P_3} = 3|g1| + 5$,
  $|unif(g, g)|_{P_3} = 0$,
  and let
  $I_{P_2} = \{rem(n, g1, g2) \mid g1, g2 \text{ lists and either } g1 = g2 = [\ ] \text{ or } |g2| < |g1|\} \cup$
  $\cup [spec(X, Y, Z, W)] \cup \{member(n, g) \mid g \text{ list and } n \text{ in } set(g)\}$,
  $I_{P_3} = [red(N1, N2, G1, G2)] \cup \{unif(x, y) \mid x = y\}$.

  It is easy to check that $I_{P_2}$ and $I_{P_3}$ are specialized models of $P_2 \ominus P_1$ and $P_3 \ominus P_2$, respectively. It remains to check the tests in points i.B and ii.

  - Consider a ground instance

$$red(n1, n2, g1, g2) \leftarrow \quad \neg unif(g1, [\ ]), spec(n1, n2, n, g1),$$
$$rem(n, g1, g), red(n1, n2, g, g2).$$

of 1). We have that:

$$|red(n1, n2, g1, g2)|_{P_3} = 3|g1| + 5 > 0 = |\neg unif(g1, [\,])|_{P_3};$$

$$|red(n1, n2, g1, g2)|_{P_3} = 3|g1| + 5 = |spec(n1, n2, n, g1)|_{P_2};$$

$$|red(n1, n2, g1, g2)|_{P_3} = 3|g1| + 5 > |g_1| + 2 = |rem(n, g1, g)|_{P_2}.$$

Now, suppose that $I_{P_2} \cup I_{P_3} \models \neg unif(g1, [\,]), rem(n, g1, g)$. Then $g$ and $g1$ are lists, $g1 \neq [\,]$, and $|g| < |g1|$. Then,

$$|red(n1, n2, g1, g2)|_{P_3} = 3|g1| + 5 > 3|g| + 5 = |red(n1, n2, g, g2)|_{P_3}.$$

– Consider a ground instance

$$red(n1, n2, g, g) \leftarrow \neg spec(n1, n2, n, g).$$

of 2). We have that:

$$|red(n1, n2, g, g)|_{P_3} = 3|g| + 5 = |spec(n1, n2, n, g)|_{P_2}.$$

Observe that the presence of the literal $\neg unif(G1, [\,])$ is fundamental to guarantee left-termination. Without it, left-termination would no longer hold (take for instance the query $red(n1, n2, [\,], g)$).

## 9. Conclusion

In this paper we proposed simple methods for proving termination of a general logic program, with respect to SLD-resolution with constructive negation and Prolog selection rule. These methods combine the notions of acceptability and acyclicity. They provide a more practical proof technique for termination, where the semantic information used is minimalized. We have illustrated the relevance of the methods by means of some examples, showing in particular that SLD-resolution augmented with Chan's constructive negation is powerful enough to formalize and implement interesting problems in non-monotonic reasoning.

We would like to conclude with an observation on related work. Apt and Pedreschi (1994) introduced a modular approach for proving acceptability of logic programs, i.e., they do not deal with programs containing negated atoms. Proving termination of *general* logic programs in a modular way, using the notion of acceptability, seems a rather difficult task, because it amounts to building a model of the completion of a program by combining models of the completions of its subprograms. Apt and Pedreschi do not tackle this problem. In this paper, we have provided an alternative way of proving termination w.r.t. the Prolog selection rule, where one tries to simplify the proof by using as little semantic information as possible, possibly in an incremental way using the methodology illustrated in Section 8.

## Acknowledgements

This research was partially supported by the Esprit Basic Research Action 6810 (Compulog 2). I would like to thank Jan Rutten for proof reading this paper, Krzysztof Apt for proposing the study of acyclic and acceptable programs, Frank Teusink for pleasant discussions, as well as the anonymous referees for useful suggestions and comments on an earlier version of this paper.

# References

Apt, K., & Bezem, M. (1991). Acyclic programs. *New Generation Computing, 9*, 335–363.

Apt, K., & Bol, R. (1994). Logic programming and negation: a survey. *The Journal of Logic Programming, 19-20*, 9–71.

Apt, K., & Pedreschi, D. (1991). Proving termination of general prolog programs. In *Proceedings of the Int. Conf. on Theoretical Aspects of Computer Software*, Vol. LNCS 526, pp. 265–289. Springer Verlag.

Baral, C., & Gelfond, M. (1994). Logic programming and knowledge representation. *The Journal of Logic Programming, 19-20*, 73–148.

Bratko, I. (1986). *PROLOG Programming for Artificial Intelligence*. Addison-Wesley.

Chan, D. (1988). Constructive negation based on the completed database. In *Proceedings of the 5th Int. Conf. and Symp. on Logic Programming*, pp. 111–125.

Clark, K. (1978). *Logic and Databases*, chap. Negation as Failure, pp. 293–322. Plenum Press, NY.

De Schreye, D., & Decorte, S. (1994). Termination of logic programs: The never-ending story. *The Journal of Logic Programming, 19-20*, 199–260.

Dershowitz, N. (1987). Termination of rewriting. *Journal of Symbolic Computation, 3*, 69–116.

Evans, C. (1990). Negation as failure as an approach to the hanks and mcdermott problem. In *Proceedings of the 2nd International Symposium om AI*, pp. 23–27.

Kowalski, R., & Sergot, M. (1986). A logic based calculus of events. *New Generation Computing, 4*, 67–95.

Marchiori, E. (1995). A methodology for proving termination of general logic programs. In *Proceedings of the 14th International Joint Conference on Artificial Intelligence (IJCAI'95)*, pp. 356–367.

Marchiori, E. (1996). On termination of general logic programs w.r.t. constructive negation. *The Journal of Logic Programming, 26(1)*, 69–89.

McCarthy, J., & Hayes, P. (1969). Some philosophical problems from the standpoint of artificial intelligence. *Machine Intelligence, 4*, 463–502.

Nilsson, N. (1982). *Principles of Artificial Intelligence*. Springer-Verlag.

Sterling, L., & Shapiro, E. (1994). *The Art of Prolog*. MIT Press.